\documentclass[letterpaper]{article} 
\usepackage{aaai2026}  
\usepackage{times}  
\usepackage{helvet}  
\usepackage{courier}  
\usepackage[hyphens]{url}  
\usepackage{graphicx} 
\usepackage{natbib}  
\usepackage{caption} 
\usepackage{algorithm}
\usepackage{algorithmic}

\usepackage{amsmath}
\usepackage{amssymb}
\usepackage{booktabs}
\usepackage{multirow}
\usepackage{makecell}
\usepackage{subcaption}
\usepackage[table]{xcolor}

\usepackage{newfloat}
\usepackage{listings}
\DeclareCaptionStyle{ruled}{labelfont=normalfont,labelsep=colon,strut=off} 
\floatstyle{ruled}
\newfloat{listing}{tb}{lst}{}
\floatname{listing}{Listing}
\title{R-Tuning: Wavelet-Decomposed Replay and Semantic Alignment for\\Continual Adaptation of Pretrained Time-Series Models}
\author {
    Tianyi Yin\textsuperscript{\rm 1,\rm 2,\rm 3},
    Jingwei Wang\textsuperscript{\rm 2}\thanks{Corresponding authors: Yunlong Ma, Jingwei Wang.},
    Chenze Wang\textsuperscript{\rm 2},
    Han Wang\textsuperscript{\rm 2},
    Jiexuan Cai\textsuperscript{\rm 2},\\
    Min Liu\textsuperscript{\rm 2},
    Yunlong Ma\textsuperscript{\rm 2}\footnotemark[1],
    Kun Gao\textsuperscript{\rm 4},
    Yuting Song\textsuperscript{\rm 3},
    Weiming Shen\textsuperscript{\rm 5}
}
\affiliations {
    \textsuperscript{\rm 1}Shanghai Research Institute for Intelligent Autonomous Systems, Tongji University\\
    \textsuperscript{\rm 2}College of Electronics and Information Engineering, Tongji University\\
    \textsuperscript{\rm 3}Institute of High Performance Computing, A*STAR\\
    \textsuperscript{\rm 4}Zhongguancun Academy\\
    \textsuperscript{\rm 5}Huazhong University of Science and Technology\\
    \{yin\_tianyi, jwwang, wangchenze, 2210126, 2431940, lmin, evanma\}@tongji.edu.cn,\\
    gaokun@bjzgca.edu.cn,
    Song\_Yuting@a-star.edu.sg,
    wshen@ieee.org
}

\usepackage{bibentry}

\begin{document}

\maketitle

\begin{abstract}
    Pre-trained models have demonstrated exceptional generalization capabilities in time-series forecasting; however, adapting them to evolving data distributions remains a significant challenge. A key hurdle lies in accessing the original training data, as fine-tuning solely on new data often leads to catastrophic forgetting. To address this issue, we propose Replay Tuning (R-Tuning), a novel framework designed for the continual adaptation of pre-trained time-series models.
    R-Tuning constructs a unified latent space that captures both prior and current task knowledge through a frequency-aware replay strategy. Specifically, it augments model-generated samples via wavelet-based decomposition across multiple frequency bands, generating trend-preserving and fusion-enhanced variants to improve representation diversity and replay efficiency. To further reduce reliance on synthetic samples, R-Tuning introduces a latent consistency constraint that aligns new representations with the prior task space. This constraint guides joint optimization within a compact and semantically coherent latent space, ensuring robust knowledge retention and adaptation.
    Extensive experimental results demonstrate the superiority of R-Tuning, which reduces MAE and MSE by up to 46.9\% and 46.8\%, respectively, on new tasks, while preserving prior knowledge with gains of up to 5.7\% and 6.0\% on old tasks. Notably, under few-shot settings, R-Tuning outperforms all state-of-the-art baselines even when synthetic proxy samples account for only 5\% of the new task dataset.
    \end{abstract}

\begin{links}
    \link{Code}{https://github.com/Ivan-YinTY/R-Tuning}
\end{links}

\section{Introduction}

Time-series forecasting plays a crucial role in various domains, including transportation, energy systems, and industrial management \cite{wang2024transformer}. Traditional forecasting methods, including statistical models and early deep learning models, have demonstrated strong performance on task-specific datasets \cite{li2023self}. However, they typically operate under a one-model-per-task paradigm, which limits their capability to diverse and evolving real-world conditions \cite{zhang2024large}.

The emergence of pre-trained models (PTMs) has enabled a shift toward general-purpose forecasting systems \cite{kim2025comprehensive}. By converting temporal signals into discrete tokens and leveraging transformer-based architectures, PTMs can perform a wide range of forecasting tasks without explicit fine-tuning \cite{gruver2023large}. These models enhance generalization and reusability across tasks. However, once released or deployed, PTMs often become static and cannot adapt smoothly to new or evolving knowledge encountered over time.

\begin{figure}[t]
\centering
\includegraphics[width=1.0\columnwidth]{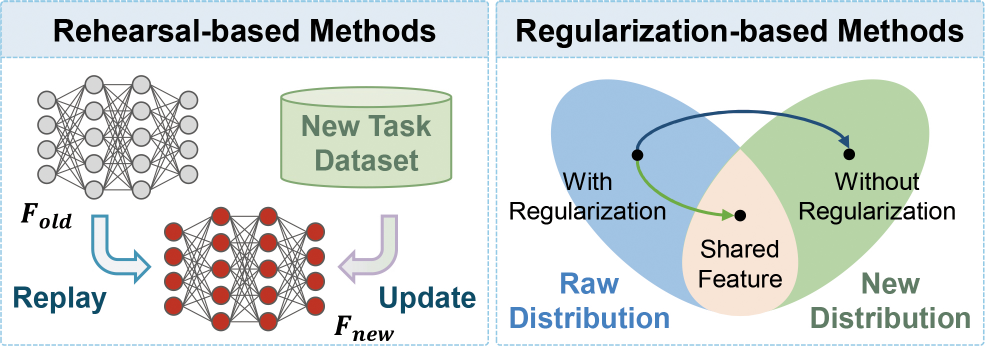} 
\caption{Two types of adaptation methods. Rehearsal-based methods mitigate forgetting by training on synthesized samples and new data. Regularization-based methods constrain parameter updates within the shared feature space.}
\label{fig:intro}
\end{figure}

Existing adaptation strategies require access to both the original training data and new task data to update the model jointly \cite{gu2024summarizing}. However, for PTMs, this is rarely practical due to concerns over data privacy, proprietary usage restrictions, and the high cost of fine-tuning large models \cite{zhou2024continual}. Conversely, updating the model with only new data causes catastrophic forgetting, where prior task performance deteriorates significantly.

Continual learning methods offer promising solutions for this issue, especially generative rehearsal and regularization-based approaches \cite{zhou2024class}. As shown in Figure~\ref{fig:intro}, generative replay synthesizes past data to retain prior knowledge, but identifying informative and representative samples from the generated distribution is non-trivial, especially for time-series signals that contain multi-scale temporal patterns \cite{cai2024msgnet}. On the other hand, regularization-based methods aim to limit the extent to which the model deviates from prior knowledge of the task. However, they often face a familiar dilemma: learn fast, forget faster—that is, learning new tasks more effectively usually means forgetting old ones more quickly \cite{biderman2024lora}.

\begin{figure*}[ht]
\centering
\includegraphics[width=1.0\textwidth]{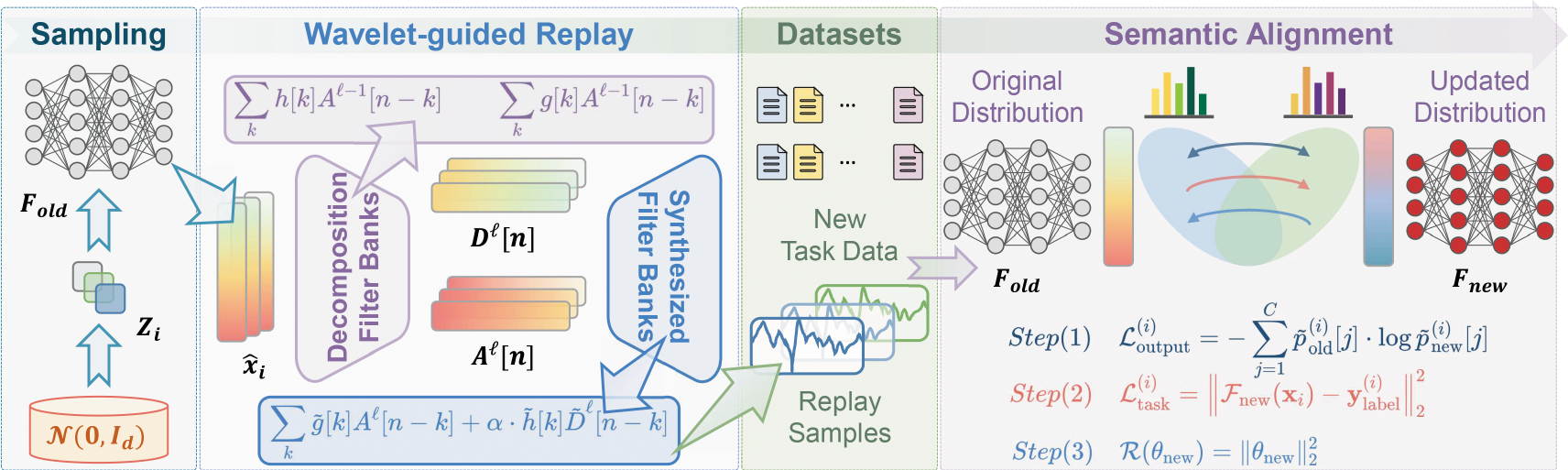} 
\caption{The architecture of R-Tuning. Random seeds are sampled from the Gaussian distribution as inputs to the old model. High-quality synthetic samples are generated via the Wavelet-guided Replay mechanism and combined with the new task dataset for fine-tuning. The Semantic Alignment mechanism ensures a performance balance between the new and old tasks.}
\label{fig:overview}
\end{figure*}

To address these challenges, we propose Replay Tuning (R-Tuning), a novel framework for continual adaptation specifically designed for PTMs in time-series forecasting. Our method integrates wavelet-based replay to improve the representativeness of old-task samples and semantic alignment to preserve latent knowledge structures across tasks, enabling stable and efficient adaptation to new tasks without forgetting. Specifically, we generate frequency-aware synthetic samples through multi-band wavelet decomposition, preserving trend-level information while enriching sample diversity. These samples are combined with new task data to guide adaptation. Simultaneously, a semantic alignment constraint ensures the latent space of the adapted model remains consistent with the original, enabling stable knowledge integration and reducing the dependence on large replay buffers.

Our contributions are summarized as follows:

\begin{enumerate}
    \item We propose R-Tuning, a continual learning method that enables efficient and stable adaptation of pre-trained time-series models through wavelet-based replay and semantic alignment.

    \item We demonstrate that using only 4\%–5\% synthetic samples (relative to the new task dataset size) is sufficient to achieve optimal performance across old and new tasks.

    \item Extensive experiments on multiple benchmarks validate the generalization and efficiency of our method, consistently outperforming state-of-the-art methods under few-shot scenarios.
\end{enumerate}

\section{Related Work}

\subsection{Pre-trained Models for Time-Series Forecasting}

Foundation models for time-series forecasting aim to generalize across diverse domains, tasks, and temporal patterns by learning transferable representations from large-scale time-series corpora \cite{liu2025timecma}. Current designs largely follow three architectural paradigms: encoder-decoder, encoder-only, and decoder-only.

\subsubsection{Encoder-Decoder Architectures}

Encoder-decoder architectures integrate a context-aware encoder with a generative decoder, enabling sequence-to-sequence modeling for forecasting tasks \cite{liu2025empowering}. These models typically support flexible conditioning and multi-resolution generation. For instance, Chronos \cite{ansari2024chronos} extends the encoder-decoder structure from the T5 language model to temporal data, introducing a task-aware forecasting interface. Apollo-Forecast \cite{yin2025apollo} further enhances Chronos by incorporating anti-aliasing tokenization and a parallel decoding mechanism, thereby improving inference accuracy and efficiency.

While encoder-decoder models exhibit strong generalization performance across a wide range of forecasting tasks, their architectural complexity leads to increased training cost and lower efficiency, especially under constrained parameter budgets.

\subsubsection{Encoder-Only Architectures}

To simplify model structure and improve training efficiency, some approaches adopt encoder-only designs that extract contextual representations from the input sequence for direct forecasting \cite{liang2024foundation}. AutoTimes and Moirai are two representative encoder-only models. AutoTimes \cite{liu2024autotimes} leverages decoder-only PTMs to perform autoregressive time-series forecasting with flexible context and minimal trainable parameters. Moirai \cite{woo2024unified} adopts a masked encoder design to address cross-frequency learning and multivariate forecasting, achieving strong zero-shot performance across diverse datasets.

However, encoder-only models suffer from limited expressiveness due to their low-rank structure \cite{dong2021attention} and often underperform in generative forecasting tasks that involve long-term, fine-grained extrapolation.

\subsubsection{Decoder-Only Architectures}

Decoder-only architectures generate time series in an autoregressive manner, akin to language models. Their unidirectional attention structure aligns naturally with the forecasting objective, enabling strong modeling of sequential dependencies \cite{das2024decoder}. Representative models include TimerXL, which utilizes linear-time attention to handle long input sequences efficiently \cite{liutimer}, and GPT4TS, which enhances generalization and forecasting accuracy by modifying the feedforward layers within residual blocks \cite{zhou2023one}. These models are favored for their architectural simplicity and long-term forecasting capabilities.

However, despite these strengths, current foundation models—regardless of design—remain static after pretraining or deployment \cite{han2025loire}. They lack mechanisms to adapt to new data distributions or evolving task requirements continually, motivating recent efforts to explore continual learning and adaptation techniques tailored for time-series foundation models.

\subsection{Continual Learning for PTMs}

Among the various continual learning paradigms, only a few are practical for PTMs due to common constraints such as fixed architectures, inaccessibility of original training data, and high retraining costs \cite{wang2024comprehensive}. In this context, two mainstream approaches have gained traction: rehearsal-based and regularization-based methods.

\subsubsection{Rehearsal-based Methods}

Rehearsal-based methods aim to preserve prior knowledge by constructing a shared feature space between old and new tasks \cite{xu2024magpie}. A common strategy involves synthesizing proxy samples from previous tasks using PTMs, then mixing them with new data during adaptation \cite{sunlamol}, enabling replay without access to the original datasets. Feature-level variants, such as RanPAC \cite{mcdonnell2023ranpac}, further reduce reliance on raw data by applying random projection to compress embeddings, followed by clustering to select a representative coreset for replay \cite{tong2025coreset}. These methods are computationally efficient and lightweight.

However, such approaches often ignore the multi-frequency nature of time-series signals. Samples selected purely based on spatial similarity may lack critical temporal structures—especially low-frequency trends and high-frequency variations—limiting the expressiveness of the replay buffer. It becomes a bottleneck in adapting PTMs for real-world forecasting tasks involving distributional shifts.

\subsubsection{Regularization-based Methods}

Regularization-based methods constrain the model to retain old knowledge by penalizing significant deviations in parameters or outputs. For example, Elastic Weight Consolidation (EWC) introduces a Fisher-based penalty to minimize updates on important weights \cite{kurniawan2024evolving}, while Learning without Forgetting (LwF) utilizes distillation to align predictions between the old and new models \cite{bonato2024mind}.

Despite their effectiveness, these methods inherently face a trade-off: restricting model drift helps preserve old tasks but often hinders learning new ones \cite{kalajdzievski2024scaling}. This “learn fast, forget faster” dilemma highlights the need for more adaptive strategies that can achieve both knowledge retention and generalization, especially in time-series settings where signals evolve over time and across domains.

\section{Methodology}

To address the challenges of adapting pre-trained time-series models without access to the original training data, we propose a novel continual fine-tuning method named Replay Tuning (R-Tuning). R-Tuning builds a compact, frequency-aware training sample space that jointly represents both the old and new task distributions. It comprises two key modules: Wavelet-guided Replay, which synthesizes and selects high-quality replay samples using frequency-aware decomposition, and Semantic Alignment, which maintains consistency between new and old task representations by aligning their latent semantic spaces. An overview of the proposed approach is shown in Figure~\ref{fig:overview}.

\begin{figure}[t]
\centering
\includegraphics[width=1.0\columnwidth]{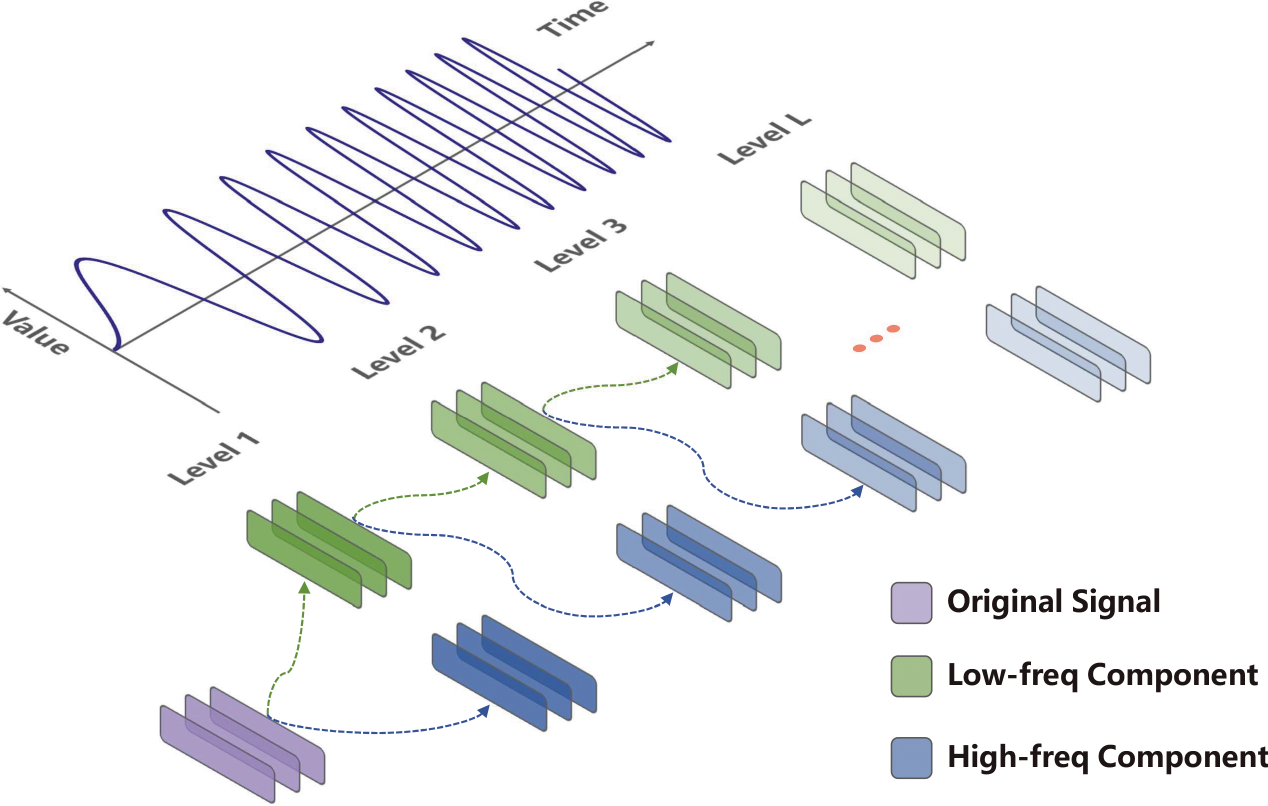} 
\caption{Decompose the signal using Redundant Wavelet Transform. Freq denotes frequency.}
\label{fig:wavelet}
\end{figure}

\subsection{Wavelet-guided Replay}
The Wavelet-guided Replay module enhances the quality and representativeness of replayed samples by leveraging the frequency structures inherent in time-series data. It consists of two components: Wavelet-based Sample Synthesis and Frequency-aware Replay Strategy.

\subsubsection{Wavelet-based Sample Synthesis}
We begin by sampling $\{\mathbf{z}_i\}_{i=1}^N$ from a standard multivariate Gaussian distribution $\mathcal{N}(\mathbf{0}, \mathbf{I})$, where $\mathbf{z}_i \in \mathbb{R}^d$ and $d$ is the input dimensionality of the pretrained model. These vectors are passed into a frozen pretrained model $\mathcal{F}_{\text{old}}$ to produce synthetic time series samples $\hat{\mathbf{x}}_i \in \mathbb{R}^T$, where $T$ denotes the sequence length. Each synthetic sequence $\hat{\mathbf{x}}_i$ captures temporal features embedded in the original training distribution, serving as proxies for previously seen data.

To extract frequency-localized components, we apply the Redundant Wavelet Transform (RWT) to each synthetic sequence. We initialize the approximation as $A^0 = \hat{\mathbf{x}}_i$, and recursively decompose the signal up to level $L$, generating a set of approximation coefficients $\{A^\ell\}_{\ell=1}^L$ and detail coefficients $\{D^\ell\}_{\ell=1}^L$ as follows:
\begin{align}
D^\ell[n] &= \sum_k h[k] A^{\ell-1}[n - k], \\
A^\ell[n] &= \sum_k g[k] A^{\ell-1}[n - k],
\end{align}
where $h[k]$ and $g[k]$ are the high-pass and low-pass decomposition filters derived from the Daubechies wavelet of order 4 (db4).

\begin{algorithm}[!h]
\caption{Replay Tuning (R-Tuning)}
\label{alg:rtuning}
\textbf{Input}: Pretrained model $\mathcal{F}_{\text{old}}$, new data $\mathcal{D}_{\text{new}}$\\
\textbf{Parameter}: Latent dim $d$, sequence length $T$, wavelet levels $L$, detail discard depth $k$, replay size $N$, scale factor $\alpha$, temperature $\tau$, loss weights $\lambda$, $\beta$\\
\textbf{Output}: Adapted model $\mathcal{F}_{\text{new}}$
\begin{algorithmic}[1]
\STATE Sample latent codes: $\{\mathbf{z}_i\}_{i=1}^{N} \sim \mathcal{N}(\mathbf{0}, \mathbf{I}_d)$
\STATE Generate synthetic samples: $\hat{\mathbf{x}}_i = \mathcal{F}_{\text{old}}(\mathbf{z}_i),\quad \hat{\mathbf{x}}_i \in \mathbb{R}^T$
\FOR{$i = 1$ to $N$}
    \STATE Initialize $A^0 = \hat{\mathbf{x}}_i$
    \FOR{$\ell = 1$ to $L$}
        \STATE $D^\ell[n] = \sum_k h[k] A^{\ell-1}[n - k]$
        \STATE $A^\ell[n] = \sum_k g[k] A^{\ell-1}[n - k]$
    \ENDFOR
    \FOR{$j = 1$ to $k$}
        \FOR{$\ell = L$ down to $1$}
            \STATE $\tilde{D}^\ell =
                \begin{cases}
                D^\ell, & \ell \leq L - j \\
                \mathbf{0}, & \ell > L - j
                \end{cases}$
            \STATE $A^{\ell-1}[n] = \sum_k \tilde{g}[k] A^{\ell}[n - k] + \alpha \cdot \tilde{h}[k] \tilde{D}^\ell[n - k]$
        \ENDFOR
        \STATE $\tilde{\mathbf{x}}_i^{(j)} = A^0$
    \ENDFOR
\ENDFOR
\STATE Construct replay set: $\mathcal{D}_{\text{replay}} = \{\hat{\mathbf{x}}_i\} \cup \{\tilde{\mathbf{x}}_i^{(j)}\}$
\STATE Merge to training set: $\mathcal{D}_{\text{train}} = \mathcal{D}_{\text{new}} \cup \mathcal{D}_{\text{replay}}$
\STATE Initialize: $\mathcal{F}_{\text{new}} \gets \mathcal{F}_{\text{old}}$
\FOR{$\mathbf{x}_i \in \mathcal{D}_{\text{train}}$}
    \STATE $\mathbf{y}_{\text{old}} = \mathcal{F}_{\text{old}}(\mathbf{x}_i), \quad \mathbf{y}_{\text{new}} = \mathcal{F}_{\text{new}}(\mathbf{x}_i)$
    \STATE $\tilde{p}_{\text{old}}^{(i)}[j] = \frac{e^{y_{\text{old},j}^{(i)}/\tau}}{\sum_{k=1}^C e^{y_{\text{old},k}^{(i)}/\tau}}, \quad \tilde{p}_{\text{new}}^{(i)}[j] = \frac{e^{y_{\text{new},j}^{(i)}/\tau}}{\sum_{k=1}^C e^{y_{\text{new},k}^{(i)}/\tau}}$
    \STATE $\mathcal{L}_{\text{output}}^{(i)} = - \sum_{j=1}^{C} \tilde{p}_{\text{old}}^{(i)}[j] \cdot \log \tilde{p}_{\text{new}}^{(i)}[j]$
    \STATE $\mathcal{L}_{\text{task}}^{(i)} = \left\| \mathcal{F}_{\text{new}}(\mathbf{x}_i) - \mathbf{y}_{\text{label}}^{(i)} \right\|_2^2$
\ENDFOR
\STATE Compute average losses:
\STATE $\mathcal{L}_{\text{output}} = \frac{1}{|\mathcal{D}_{\text{train}}|} \sum_i \mathcal{L}_{\text{output}}^{(i)}$
\STATE $\mathcal{L}_{\text{task}} = \frac{1}{|\mathcal{D}_{\text{train}}|} \sum_i \mathcal{L}_{\text{task}}^{(i)}$
\STATE Compute regularization: $\mathcal{R}(\theta_{\text{new}}) = \|\theta_{\text{new}}\|_2^2$
\STATE Compute total loss:
\STATE $\mathcal{L}_{\text{total}} = \mathcal{L}_{\text{task}} + \lambda \mathcal{L}_{\text{output}} + \beta \cdot \mathcal{R}(\theta_{\text{new}})$
\STATE Update $\mathcal{F}_{\text{new}}$ via gradient descent on $\mathcal{L}_{\text{total}}$
\STATE \textbf{return} $\mathcal{F}_{\text{new}}$
\end{algorithmic}
\end{algorithm}

The above process is shown in Figure~\ref{fig:wavelet}. The decomposition low-pass filter coefficients for db4 are expressed as:
\begin{align}
g[0] &= \frac{1 + \sqrt{3}}{4\sqrt{2}}, & g[1] &= \frac{3 + \sqrt{3}}{4\sqrt{2}}, \\
g[2] &= \frac{3 - \sqrt{3}}{4\sqrt{2}}, & g[3] &= \frac{1 - \sqrt{3}}{4\sqrt{2}},
\end{align}
and the corresponding high-pass decomposition filter is defined by:
\begin{equation}
h[k] = (-1)^k g[3 - k], \quad \text{for } k = 0, 1, 2, 3.
\end{equation}

The reconstruction filters $\tilde{g}[k]$ and $\tilde{h}[k]$ are derived using time-reversal and sign alternation operations on $g[k]$ and $h[k]$, respectively:
\begin{equation}
\tilde{g}[k] = g[k], \quad \tilde{h}[k] = (-1)^k g[k],
\end{equation}
forming the synthesis filter bank used to reconstruct time-domain signals from wavelet coefficients and ensuring perfect reconstruction.

Unlike the discrete wavelet transform, which incorporates downsampling that may introduce aliasing even with orthogonal mirror filters, the RWT avoids this issue by inserting zeros between coefficients. This design enhances the effective receptive field while preserving temporal resolution.

\subsubsection{Frequency-aware Replay Strategy}
Building upon the multiscale decomposition described above, we construct multiple variants of each synthetic signal, each emphasizing distinct spectral characteristics to reflect specific temporal properties. Specifically, we remove the highest $k$ levels of detail coefficients $\{D^{L-k+1}, \ldots, D^L\}$ and perform reconstruction via the Inverse Redundant Wavelet Transform.

For each level $\ell$ from $L$ down to $1$, the reconstruction is carried out iteratively as:
\begin{equation}
A^{\ell-1}[n] = \sum_k \tilde{g}[k] A^{\ell}[n - k] + \alpha \cdot \tilde{h}[k] D^{\ell}[n - k],
\end{equation}
where $\alpha \in [0, 1]$ is a level-dependent detail scaling coefficient that controls the contribution of high-frequency components during reconstruction. This modification enables smoother spectral filtering and generates frequency-aware replay samples with varying degrees of local variation. After completing all inverse steps, we obtain the reconstructed signal $\tilde{\mathbf{x}}_i^{(L-k)} = A^0$, which preserves components from the lower-frequency subbands while omitting selected high-frequency details.

This strategy enables the generation of multiple replay samples for each synthetic sequence, each emphasizing different spectral bands and encoding complementary characteristics, such as global trends and local fluctuations. These frequency-aware variants enrich the replay set by expanding the diversity of temporal dynamics while preserving task-specific features from previous domains.

We further construct the final training set $\mathcal{D}_{\text{train}}$ by combining the new task samples with the full-spectrum and frequency-filtered synthetic signals:
\begin{equation}
\mathcal{D}_{\text{train}} = \mathcal{D}_{\text{new}} \cup \{\hat{\mathbf{x}}_i\}_{i=1}^N \cup \{\tilde{\mathbf{x}}_i^{(L-k)}\}_{i=1}^{N'}.
\end{equation}

This comprehensive dataset facilitates continual adaptation by simultaneously encoding old-task knowledge and new-task representations.

\subsection{Semantic Alignment via Latent Distillation}
To complement the Wavelet-guided Replay module and further improve the retention of prior knowledge while reducing reliance on synthetic sample quality, we introduce a Semantic Alignment module based on latent space distillation. It enhances the consistency between the new model's output distribution and that of the raw pretrained model, thereby reinforcing task-specific knowledge and promoting more stable continual adaptation without requiring access to original training data.

\subsubsection{Distillation Mechanism}  
To mitigate semantic drift and prevent catastrophic forgetting during continual adaptation, we adopt a distillation mechanism that transfers knowledge from the original pretrained model $\mathcal{F}_{\text{old}}$ to the new model $\mathcal{F}_{\text{new}}$. This mechanism aligns the output distributions of the two models on the replay-augmented dataset $\mathcal{D}_{\text{train}}$, thereby maintaining consistency with prior knowledge even when original task labels are unavailable.

For each training input $\mathbf{x}_i \in \mathcal{D}_{\text{train}}$, we obtain the output logits from both models:
\begin{equation}
\mathbf{y}_{\text{new}} = \mathcal{F}_{\text{new}}(\mathbf{x}_i), \quad 
\mathbf{y}_{\text{old}} = \mathcal{F}_{\text{old}}(\mathbf{x}_i).
\end{equation}

To produce softened prediction distributions that reveal relative confidence, we apply temperature scaling with $\tau \in (0, 1]$ directly to the logits. Specifically, the softened prediction probabilities for class $j$ are given by:
\begin{equation}
\tilde{p}_{\text{old}}^{(i)}[j] = \frac{\exp(y_{\text{old},j}^{(i)} / \tau)}{\sum_{k=1}^{C} \exp(y_{\text{old},k}^{(i)} / \tau)}, 
\end{equation}
\begin{equation}
\tilde{p}_{\text{new}}^{(i)}[j] = \frac{\exp(y_{\text{new},j}^{(i)} / \tau)}{\sum_{k=1}^{C} \exp(y_{\text{new},k}^{(i)} / \tau)}, 
\end{equation}
where $C$ is the number of output classes and $y_{\text{old},j}^{(i)}$ and $y_{\text{new},j}^{(i)}$ are the $j$-th logits from the old and new models, respectively. The distillation loss is computed as the cross-entropy between these softened distributions:
\begin{equation}
\mathcal{L}_{\text{output}}^{(i)} = - \sum_{j=1}^{C} \tilde{p}_{\text{old}}^{(i)}[j] \cdot \log \tilde{p}_{\text{new}}^{(i)}[j],
\end{equation}
and averaged across all training samples:
\begin{equation}
\mathcal{L}_{\text{output}} = \frac{1}{|\mathcal{D}_{\text{train}}|} \sum_{\mathbf{x}_i \in \mathcal{D}_{\text{train}}} \mathcal{L}_{\text{output}}^{(i)}.
\end{equation}

This loss encourages $\mathcal{F}_{\text{new}}$ to approximate the predictive behavior of $\mathcal{F}_{\text{old}}$, preserving prior task knowledge under a softened output distribution. Temperature scaling not only improves the informativeness of supervision by spreading the probability mass but also smooths the optimization landscape. From a gradient perspective, the derivative of the temperature-scaled softmax is:
\begin{equation}
\frac{\partial \tilde{p}[j]}{\partial z_m} = \frac{1}{\tau} \cdot \tilde{p}[j] \cdot (\delta_{jm} - \tilde{p}[m]),
\end{equation}
where $\tilde{p}[j]$ denotes the temperature-scaled probability and $\delta_{jm}$ is the Kronecker delta.

As $\tau$ increases, the gradient magnitude decreases, which flattens the loss surface and stabilizes training. Therefore, distillation not only acts as a semantic regularizer but also as an optimization stabilizer for continual learning.

\subsubsection{Training Objective}
The training objective integrates both the original task loss and the distillation loss. On one hand, we minimize the task-specific loss using the ground truth targets:
\begin{equation}
\mathcal{L}_{\text{task}} = \frac{1}{|\mathcal{D}_{\text{train}}|} \sum_{\mathbf{x}_i \in \mathcal{D}_{\text{train}}} \left\| \mathcal{F}_{\text{new}}(\mathbf{x}_i) - \mathbf{y}_{\text{label}}^{(i)} \right\|_2^2,
\end{equation}
where $\mathbf{y}_{\text{label}}^{(i)}$ is the ground truth target for input $\mathbf{x}_i$.

At the same time, we incorporate the knowledge from the frozen model via the distillation loss, leading to a composite objective:
\begin{equation}
\mathcal{L}_{\text{total}} = \mathcal{L}_{\text{task}} + \lambda \mathcal{L}_{\text{output}} + \beta \mathcal{R}(\theta),
\end{equation}
where $\lambda$ and $\beta$ are hyperparameters that control the trade-off among the task loss, the distillation loss, and a regularization term $\mathcal{R}(\theta)$ is implemented via L2 weight decay in the Adam.

This joint optimization enables the model to integrate new knowledge while maintaining performance on previous tasks, thereby improving stability and generalization in continual learning.

\begin{table*}[tb]
\centering
\small
\renewcommand{\arraystretch}{1.1}
\begin{tabular}{@{}cc|l|l|l|l|l|l@{}}
\toprule
& \multicolumn{1}{c|}{\textbf{Method}} 
& \multicolumn{1}{c|}{\textbf{Chronos}} 
& \multicolumn{1}{c|}{\textbf{Apollo}} 
& \multicolumn{1}{c|}{\textbf{Moirai}} 
& \multicolumn{1}{c|}{\textbf{AutoTimes}} 
& \multicolumn{1}{c|}{\textbf{TimerXL}} 
& \multicolumn{1}{c@{}}{\textbf{GPT4TS}} \\ 
\midrule
\multirow{8}{*}{\rotatebox{90}{\textbf{MAE}}}
& \multicolumn{1}{c|}{Raw}     
& \multicolumn{1}{c|}{5.3$\pm$0.7} & \multicolumn{1}{c|}{5.3$\pm$1.0} & \multicolumn{1}{c|}{5.2$\pm$1.1} & \multicolumn{1}{c|}{5.5$\pm$1.1} & \multicolumn{1}{c|}{4.9$\pm$0.9} & \multicolumn{1}{c}{5.6$\pm$1.0} \\
\cmidrule(lr){2-8}
& FT      
& 5.4$\pm$0.8/-1.13\% 
& 5.3$\pm$0.6/-0.19\% 
& 5.3$\pm$1.2/-2.31\% 
& 5.6$\pm$1.5/-2.56\% 
& 4.9$\pm$0.3/+1.21\% 
& 5.6$\pm$1.0/+0.71\% \\
& Frozen  
& 5.3$\pm$0.4/-0.75\% 
& 5.3$\pm$1.3/+0.00\% 
& 5.1$\pm$0.7/+1.92\% 
& 5.4$\pm$0.9/+0.92\% 
& 5.0$\pm$1.1/-1.42\% 
& 5.5$\pm$1.4/+2.32\% \\
& LwF     
& 5.3$\pm$0.5/+0.19\% 
& 5.2$\pm$1.0/+1.51\% 
& \underline{5.0$\pm$0.8/+3.27\%}
& 5.5$\pm$1.2/-1.10\% 
& 4.8$\pm$0.7/+2.63\% 
& 5.4$\pm$0.6/+3.57\% \\
& EWC     
& 5.2$\pm$1.4/+1.13\% 
& 5.2$\pm$0.9/+1.13\% 
& 5.2$\pm$1.1/-0.77\% 
& 5.3$\pm$1.3/+2.38\% 
& 4.9$\pm$0.5/+0.61\% 
& 5.5$\pm$0.8/+1.60\% \\
& RanPAC  
& 5.2$\pm$0.3/+1.51\% 
& 5.1$\pm$0.7/+3.77\% 
& 5.1$\pm$1.5/+1.73\% 
& \underline{5.1$\pm$0.4/+6.59\%}
& 4.8$\pm$0.9/+2.23\% 
& 5.5$\pm$1.2/+1.96\% \\
& LAMOL   
& 5.3$\pm$0.6/+0.00\% 
& 5.2$\pm$1.4/+1.13\% 
& 5.1$\pm$0.5/+1.35\% 
& 5.4$\pm$1.0/+0.55\% 
& \underline{4.8$\pm$1.3/+3.04\%} 
& 5.4$\pm$0.7/+3.03\% \\
\cmidrule(l){2-8}
\rowcolor{black!5}
& \textbf{R-Tuning} 
& \underline{5.2$\pm$1.1/+2.83\%} 
& \underline{5.1$\pm$0.8/+4.15\%} 
& 5.1$\pm$1.3/+2.31\%
& 5.2$\pm$0.6/+4.21\%
& 4.9$\pm$0.2/+1.62\%
& \underline{5.3$\pm$1.5/+5.70\%} \\
\midrule
\multirow{8}{*}{\rotatebox{90}{\textbf{MSE}}}
& \multicolumn{1}{c|}{Raw}     
& \multicolumn{1}{c|}{6.9$\pm$1.1} & \multicolumn{1}{c|}{6.9$\pm$1.0} & \multicolumn{1}{c|}{6.7$\pm$0.9} & \multicolumn{1}{c|}{7.1$\pm$0.8} & \multicolumn{1}{c|}{6.4$\pm$1.1} & \multicolumn{1}{c}{7.3$\pm$1.1} \\
\cmidrule(lr){2-8}
& FT      
& 7.4$\pm$1.3/-8.31\% 
& 7.8$\pm$0.4/-13.70\% 
& 7.4$\pm$0.7/-9.36\% 
& 7.7$\pm$1.0/-8.78\% 
& 7.0$\pm$1.5/-9.08\% 
& 8.1$\pm$0.9/-11.85\% \\
& Frozen  
& \underline{6.6$\pm$0.6/+3.21\%}
& 6.8$\pm$1.1/+0.73\% 
& 6.7$\pm$0.3/+0.00\% 
& 6.9$\pm$0.8/+1.56\% 
& 6.4$\pm$1.2/-0.47\% 
& 7.5$\pm$0.7/-2.89\% \\
& LwF     
& 6.9$\pm$1.0/-0.87\% 
& 7.0$\pm$1.5/-2.19\% 
& 6.8$\pm$1.4/-0.59\% 
& 7.1$\pm$0.5/-0.71\% 
& 6.5$\pm$0.4/-2.19\% 
& 7.2$\pm$1.3/+0.69\% \\
& EWC     
& 7.2$\pm$1.2/-4.37\% 
& 7.1$\pm$0.7/-3.35\% 
& 6.8$\pm$0.9/-0.89\% 
& 7.1$\pm$0.6/-1.27\% 
& 6.5$\pm$1.0/-2.19\% 
& 7.4$\pm$1.4/-1.93\% \\
& RanPAC  
& 6.7$\pm$0.5/+3.06\% 
& 6.8$\pm$0.8/+0.29\% 
& 6.4$\pm$1.1/+4.31\% 
& \underline{6.8$\pm$0.7/+3.97\%}
& 6.2$\pm$0.9/+2.82\% 
& \underline{6.9$\pm$1.2/+4.41\%} \\
& LAMOL   
& 6.9$\pm$1.4/+0.15\% 
& 7.0$\pm$0.3/-2.04\% 
& 6.6$\pm$0.8/+2.23\% 
& 6.9$\pm$1.1/+2.83\% 
& 6.4$\pm$0.6/+0.00\% 
& 7.0$\pm$0.5/+3.44\% \\
\cmidrule(l){2-8}
\rowcolor{black!5}
& \textbf{R-Tuning}
& 6.7$\pm$1.0/+2.77\%
& \underline{6.5$\pm$1.5/+5.98\%} 
& \underline{6.4$\pm$0.4/+5.50\%} 
& 6.8$\pm$0.9/+3.82\%
& \underline{6.1$\pm$0.7/+4.23\%} 
& 7.1$\pm$0.3/+2.07\% \\
\bottomrule
\end{tabular}
\caption{Average MAE/MSE (with error margins) on \textbf{Old} tasks. The relative percentages (calculated with 3 decimal places precision) are provided after slashes. Best results are underlined. All absolute values have been scaled by a factor of 10.}
\label{tab:comparison-old}
\end{table*}

\begin{table*}[!h]
\centering
\small
\setlength{\tabcolsep}{1.7mm}
\renewcommand{\arraystretch}{1.1}
\begin{tabular}{@{}cc|l|l|l|l|l|l@{}}
\toprule
& \multicolumn{1}{c|}{\textbf{Method}} 
& \multicolumn{1}{c|}{\textbf{Chronos}} 
& \multicolumn{1}{c|}{\textbf{Apollo}} 
& \multicolumn{1}{c|}{\textbf{Moirai}} 
& \multicolumn{1}{c|}{\textbf{AutoTimes}} 
& \multicolumn{1}{c|}{\textbf{TimerXL}} 
& \multicolumn{1}{c@{}}{\textbf{GPT4TS}} \\ 
\midrule
\multirow{8}{*}{\rotatebox{90}{\textbf{MAE}}}
& \multicolumn{1}{c|}{Raw}     
& \multicolumn{1}{c|}{1.9$\pm$0.3} & \multicolumn{1}{c|}{1.9$\pm$0.2} & \multicolumn{1}{c|}{1.9$\pm$0.4} & \multicolumn{1}{c|}{2.0$\pm$0.3} & \multicolumn{1}{c|}{1.8$\pm$0.3} & \multicolumn{1}{c}{2.0$\pm$0.2} \\
\cmidrule(lr){2-8}
& FT      
& 1.1$\pm$0.2/+42.19\% 
& 1.1$\pm$0.3/+44.27\% 
& 1.1$\pm$0.1/+41.27\% 
& 1.1$\pm$0.4/+43.43\% 
& 1.0$\pm$0.3/+41.90\% 
& 1.2$\pm$0.2/+42.65\% \\
& Frozen  
& 1.9$\pm$0.3/+2.08\% 
& 1.9$\pm$0.1/+3.12\% 
& 1.9$\pm$0.2/+1.59\% 
& 2.0$\pm$0.4/-0.51\% 
& 1.7$\pm$0.2/+3.35\% 
& 2.0$\pm$0.3/+4.41\% \\
& LwF     
& 1.3$\pm$0.4/+33.33\% 
& 1.2$\pm$0.2/+34.90\% 
& 1.2$\pm$0.3/+35.98\% 
& 1.3$\pm$0.1/+34.85\% 
& 1.2$\pm$0.2/+33.52\% 
& 1.4$\pm$0.4/+33.82\% \\
& EWC     
& 1.2$\pm$0.1/+35.42\% 
& 1.2$\pm$0.3/+38.02\% 
& 1.2$\pm$0.2/+35.98\% 
& 1.3$\pm$0.2/+32.32\% 
& 1.1$\pm$0.4/+36.31\% 
& 1.3$\pm$0.1/+35.29\% \\
& RanPAC  
& 1.2$\pm$0.3/+39.58\% 
& 1.1$\pm$0.2/+42.71\% 
& 1.1$\pm$0.1/+41.27\% 
& 1.1$\pm$0.3/+42.42\% 
& 1.0$\pm$0.2/+41.90\% 
& 1.2$\pm$0.4/+41.18\% \\
& LAMOL   
& \underline{1.0$\pm$0.3/+45.31\%} 
& 1.1$\pm$0.2/+41.67\% 
& \underline{1.1$\pm$0.4/+43.92\%} 
& \underline{1.1$\pm$0.1/+44.95\%}
& \underline{1.0$\pm$0.2/+43.02\%}
& 1.2$\pm$0.3/+43.14\% \\
\cmidrule(l){2-8}
\rowcolor{black!5}
& \textbf{R-Tuning} 
& \underline{1.0$\pm$0.2/+45.31\%} 
& \underline{1.0$\pm$0.3/+46.88\%} 
& 1.1$\pm$0.1/+41.27\%
& 1.1$\pm$0.4/+44.44\%
& 1.0$\pm$0.3/+41.90\%
& \underline{1.1$\pm$0.2/+46.57\%} \\
\midrule
\multirow{8}{*}{\rotatebox{90}{\textbf{MSE}}}
& \multicolumn{1}{c|}{Raw}     
& \multicolumn{1}{c|}{1.5$\pm$0.2} & \multicolumn{1}{c|}{1.5$\pm$0.3} & \multicolumn{1}{c|}{1.5$\pm$0.1} & \multicolumn{1}{c|}{1.6$\pm$0.4} & \multicolumn{1}{c|}{1.4$\pm$0.2} & \multicolumn{1}{c}{1.6$\pm$0.3} \\
\cmidrule(lr){2-8}
& FT      
& \underline{0.9$\pm$0.3/+42.21\%} 
& 0.9$\pm$0.1/+44.16\% 
& 0.9$\pm$0.2/+39.07\% 
& 1.0$\pm$0.2/+39.24\% 
& \underline{0.8$\pm$0.4/+42.66\%}
& 1.0$\pm$0.1/+41.72\% \\
& Frozen  
& 1.5$\pm$0.4/+3.90\% 
& 1.4$\pm$0.2/+6.49\% 
& 1.4$\pm$0.3/+5.30\% 
& 1.5$\pm$0.1/+5.70\% 
& 1.4$\pm$0.3/+5.59\% 
& 1.5$\pm$0.2/+4.91\% \\
& LwF     
& 1.0$\pm$0.1/+35.71\% 
& 0.9$\pm$0.4/+39.61\% 
& 0.9$\pm$0.3/+40.40\% 
& 1.0$\pm$0.2/+38.61\% 
& 0.8$\pm$0.1/+41.96\% 
& 1.0$\pm$0.4/+36.81\% \\
& EWC     
& 0.9$\pm$0.2/+38.96\% 
& 0.9$\pm$0.3/+40.91\% 
& 0.9$\pm$0.1/+39.74\% 
& 1.0$\pm$0.3/+39.87\% 
& 0.9$\pm$0.2/+39.16\% 
& 1.0$\pm$0.4/+38.04\% \\
& RanPAC  
& 1.0$\pm$0.3/+37.66\% 
& 0.9$\pm$0.1/+42.86\% 
& 0.9$\pm$0.2/+42.38\% 
& 0.9$\pm$0.4/+40.51\% 
& 0.8$\pm$0.3/+41.96\% 
& 0.9$\pm$0.2/+42.33\% \\
& LAMOL   
& 0.9$\pm$0.4/+40.26\% 
& 0.9$\pm$0.2/+40.91\% 
& 0.9$\pm$0.1/+41.72\% 
& 0.9$\pm$0.3/+40.51\% 
& 0.9$\pm$0.2/+39.86\% 
& \underline{0.9$\pm$0.3/+44.17\%} \\
\cmidrule(l){2-8}
\rowcolor{black!5}
& \textbf{R-Tuning}
& \underline{0.9$\pm$0.1/+42.21\%} 
& \underline{0.8$\pm$0.2/+46.75\%} 
& \underline{0.9$\pm$0.3/+43.71\%} 
& \underline{0.9$\pm$0.4/+42.41\%} 
& 0.8$\pm$0.2/+41.26\%
& 0.9$\pm$0.1/+43.56\% \\
\bottomrule
\end{tabular}
\caption{Average MAE/MSE (with error margins) on \textbf{New} tasks. The relative percentages (calculated with 3 decimal places precision) are provided after slashes. Best results are underlined. All absolute values have been scaled by a factor of 10.}
\label{tab:comparison-new}
\end{table*}

\section{Experiments}

\begin{figure*}[ht]
\centering
\begin{subfigure}{0.235\textwidth}
    \centering
    \includegraphics[width=\textwidth]{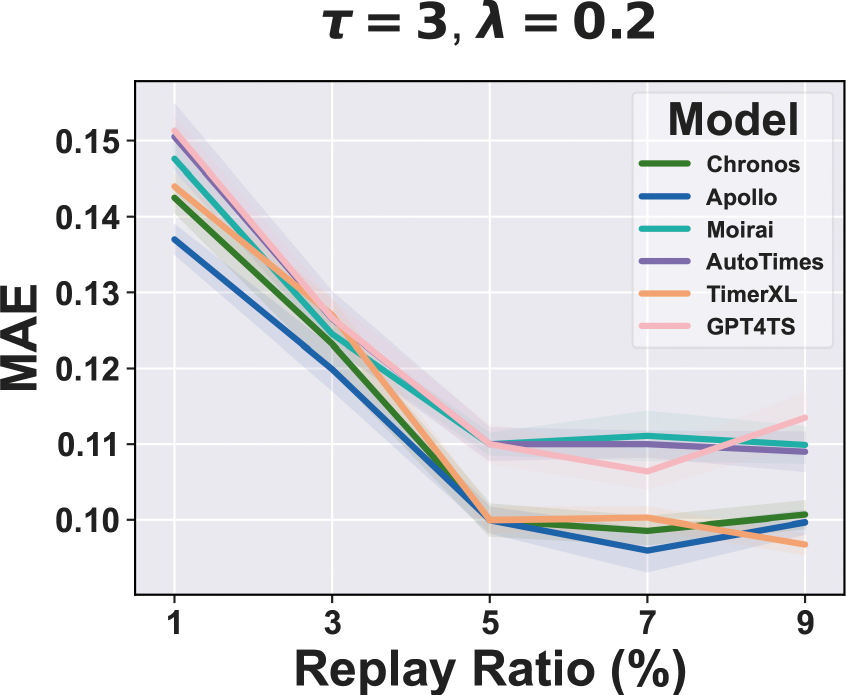}
\end{subfigure}
\hspace{0.01\textwidth}
\begin{subfigure}{0.235\textwidth}
    \centering
    \includegraphics[width=\textwidth]{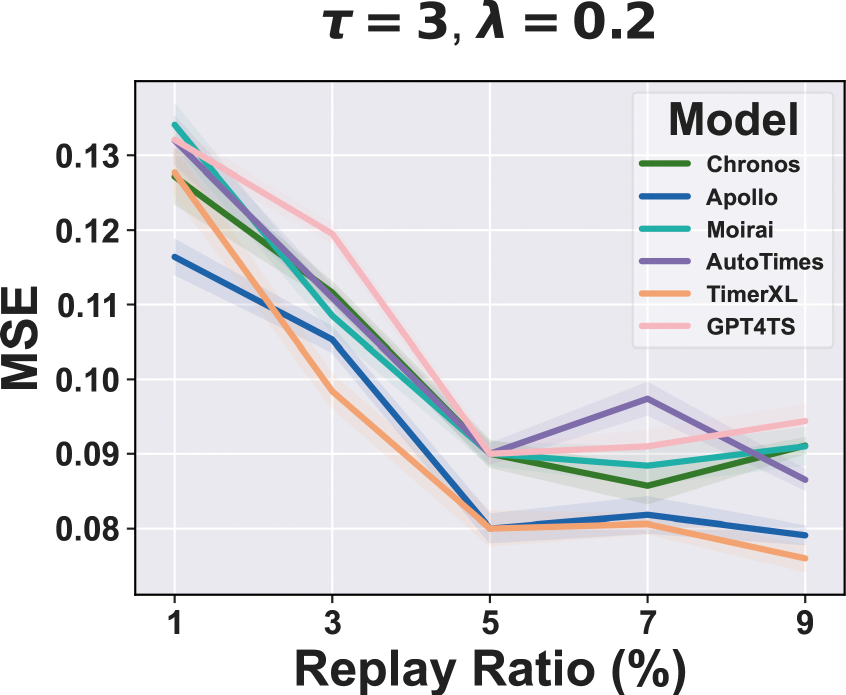}
\end{subfigure}
\hspace{0.01\textwidth}
\begin{subfigure}{0.235\textwidth}
    \centering
    \includegraphics[width=\textwidth]{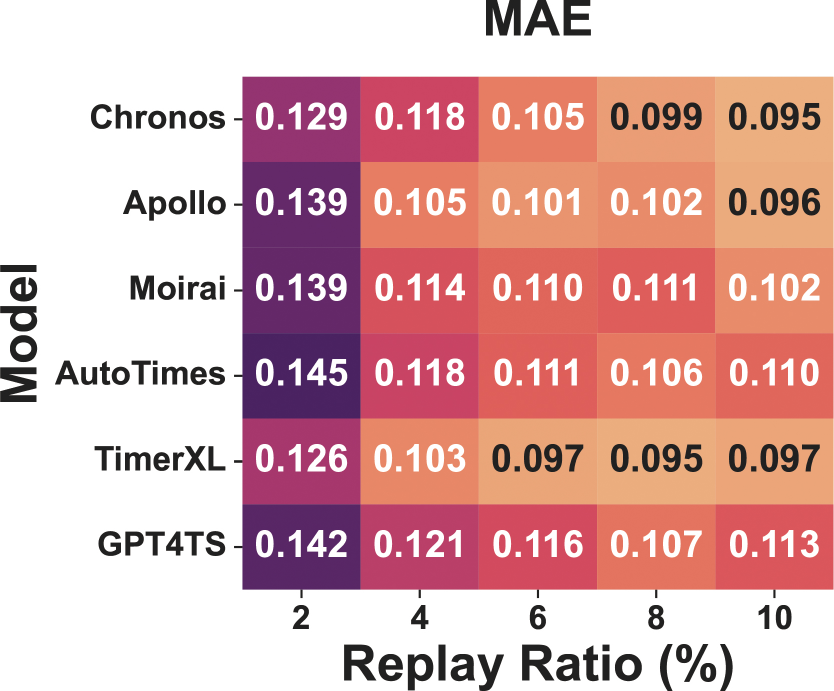}
\end{subfigure}
\hspace{0.01\textwidth}
\begin{subfigure}{0.235\textwidth}
    \centering
    \includegraphics[width=\textwidth]{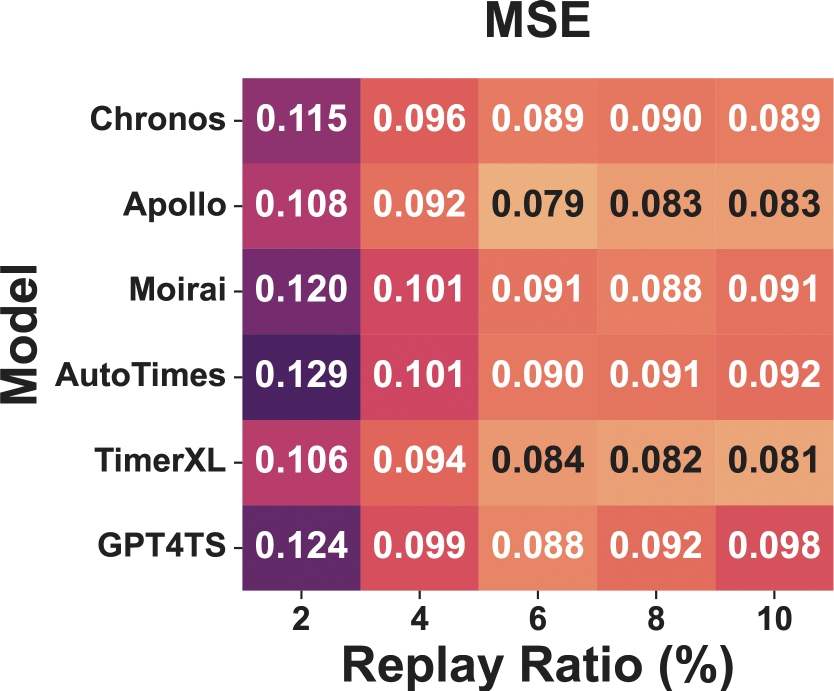}
\end{subfigure}
\caption{The performance under different proportions of the synthetic replay sample.}
\label{fig:EXP2}
\end{figure*}

\subsection{Baselines and Experimental Settings}
To assess the effectiveness of the proposed method, we conduct experiments using six SOTA pretrained time-series forecasting models, spanning encoder-decoder (Chronos, Apollo), encoder-only (Moirai, AutoTimes), and decoder-only (TimerXL, GPT4TS) architectures. These models represent diverse architectural designs and are widely adopted as strong baselines in time-series prediction tasks. We compare our approach with four categories of continual adaptation methods: rehearsal-based (RanPAC, LAMOL), regularization-based (LwF, EWC), vanilla fine-tuning (FT), and frozen-parameter adaptation. Experiments are conducted on five real-world public datasets—ETT, ECL, Traffic, Weather, and Solar—covering a broad range of domains, including energy, meteorology, and transportation \cite{liu2024timer}. To further avoid potential data leakage issues during pretraining, we also include an in-house crowd flow dataset collected from a commercial area in Shanghai, spanning from January 1, 2024, to July 1, 2025, with pedestrian counts recorded at 5-minute intervals.  All experiments are run on cloud servers equipped with dual L40 (48GB) GPUs.

\subsection{Main Results}
\subsubsection{Performance Comparison}
To evaluate the effectiveness of R-Tuning, we compare it against six representative continual adaptation methods, including rehearsal-based, regularization-based, vanilla fine-tuning, and parameter-freezing approaches. All experiments are conducted using official pretrained time-series forecasting models.

In our experimental configuration, the five public datasets are designated as old tasks, while an in-house pedestrian flow dataset is used as the new task. Since these public datasets are frequently used in the pretraining of existing models, we assume the base models have already been extensively exposed to them. Meanwhile, to mitigate the impact of varying initial performances across base models and datasets, we calculate the average MAE and MSE over the five old tasks. This strategy ensures a more reliable assessment of knowledge retention for old tasks. Each model is then adapted to the new task using only 10\% of its training set under a few-shot setting. R-Tuning further incorporates 2000 synthetic replay samples via Wavelet-guided Replay and applies latent distillation with $\tau = 3$ and $\lambda = 0.2$.

As shown in Table~\ref{tab:comparison-old} and \ref{tab:comparison-new}, R-Tuning achieves consistent performance improvements across diverse architectures and task settings. In the best case, it reduces new-task MAE and MSE by 46.88\% and 46.75\%, respectively (using Apollo), while simultaneously improving old-task MAE and MSE by up to 5.70\% (GPT4TS) and 5.98\% (Apollo), respectively. This result underscores not only our framework's enhanced adaptability to novel tasks but also its superior retention of prior knowledge and robust stability across previous tasks. Compared with baseline methods, R-Tuning consistently outperforms alternatives, offering average gains of approximately 2–5\% on old tasks and over 40\% on new tasks across all models (see the Appendix for more details).

\begin{table}[tb]
\centering
\small
\renewcommand{\arraystretch}{1.1}
\begin{tabular}{@{}cc|c|c@{}}
\toprule
& \textbf{Method} 
& \textbf{Old} 
& \textbf{New} \\
\midrule
\multirow{5}{*}{\rotatebox{90}{\textbf{MAE}}}
& Raw                 
& $0.55 \pm 0.11$ 
& $0.20 \pm 0.03$ \\
\cmidrule(lr){2-4}
& Vanilla FT         
& $0.56 \pm 0.15$ 
& $0.11 \pm 0.04$ \\
& R-Tuning w/o Replay
& $0.55 \pm 0.12$ 
& $0.13 \pm 0.01$ \\
& R-Tuning w/o Align 
& $0.54 \pm 0.10$ 
& $0.11 \pm 0.01$ \\
\rowcolor{black!5}
& \textbf{R-Tuning}     
& \textbf{$0.52 \pm 0.06$} 
& \textbf{$0.11 \pm 0.04$} \\
\midrule
\multirow{5}{*}{\rotatebox{90}{\textbf{MSE}}}
& Raw                 
& $0.71 \pm 0.08$ 
& $0.16 \pm 0.04$ \\
\cmidrule(lr){2-4}
& Vanilla FT         
& $0.77 \pm 0.10$ 
& $0.10 \pm 0.02$ \\
& R-Tuning w/o Replay
& $0.71 \pm 0.05$ 
& $0.10 \pm 0.02$ \\
& R-Tuning w/o Align 
& $0.69 \pm 0.11$ 
& $0.09 \pm 0.03$ \\
\rowcolor{black!5}
& \textbf{R-Tuning}     
& \textbf{$0.68 \pm 0.09$} 
& \textbf{$0.09 \pm 0.04$} \\
\bottomrule
\end{tabular}
\caption{Ablation results on AutoTimes using 2000 synthetic samples and latent distillation ($\tau = 3$, $\lambda = 0.2$). Best results are highlighted.}
\label{tab:ablation}
\end{table}

\subsubsection{Replay Sample Efficiency}
To investigate the impact of the number of replay samples on continual adaptation performance, we conduct a controlled experiment that focuses on varying the quantity of replay samples. Specifically, we adjust the ratio of synthetic replay samples from 1\% to 10\% of the new-task training set using the Wavelet-guided Replay strategy. Except for that, all other settings are kept consistent with those in the previous section.

As illustrated in Figure~\ref{fig:EXP2}, model performance consistently improves as the proportion of synthetic replay data increases from 1\% to 5\%, with the average MAE decreasing from 0.1455 to 0.1050, and the average MSE dropping from 0.1283 to 0.0867, across all evaluated models. Beyond a replay ratio of 5\%, however, the performance gains plateau, with MAE and MSE stabilizing around 0.10 and 0.08, respectively. A similar trend is observed in the corresponding heatmaps: as the replay ratio increases, the color intensity gradually fades and eventually stabilizes. These results suggest that R-Tuning achieves near-optimal performance with as little as 4\%–5\% synthetic replay data, underscoring the effectiveness of its frequency-aware augmentation strategy in low-replay regimes.

\subsubsection{Ablation Study}

To disentangle the contributions of key components in R-Tuning, we conduct an ablation study comparing four configurations: (1) fine-tuning without any continual learning methods; (2) R-Tuning without Wavelet-guided Replay; (3) R-Tuning without Semantic Alignment; and (4) full R-Tuning with both components enabled. All experiments follow the same protocol and evaluation metrics as in the \textit{Performance Comparison} section. We use AutoTimes as the base model in the following experiments.

Table~\ref{tab:ablation} reports the ablation results. For old tasks, vanilla fine-tuning increases the MAE from $0.55 \pm 0.11$ to $0.56 \pm 0.15$ and the MSE from $0.71 \pm 0.08$ to $0.77 \pm 0.10$, indicating clear forgetting. Introducing Semantic Alignment alone reduces the MAE and MSE to $0.55 \pm 0.12$ and $0.71 \pm 0.05$, while Wavelet-guided Replay alone achieves $0.54 \pm 0.10$ and $0.69 \pm 0.11$. The full R-Tuning configuration achieves the best retention, with MAE and MSE further reduced to $0.52 \pm 0.06$ and $0.68 \pm 0.09$. On new tasks, all configurations outperform the pretrained model ($0.20 \pm 0.03$, $0.16 \pm 0.04$), and the full setup maintains strong adaptation performance ($0.11 \pm 0.04$, $0.09 \pm 0.04$).


These results demonstrate that Semantic Alignment mitigates semantic drift by stabilizing latent representations, while Wavelet-guided Replay enriches the training distribution through frequency-aware synthesis. Their combination yields the best balance between knowledge retention and new-task adaptation.

\section{Conclusion}



In this paper, we propose R-Tuning, a replay-enhanced continual adaptation framework that integrates Wavelet-guided Replay and Semantic Alignment to balance knowledge retention and adaptation. It outperforms SOTA baselines with minimal synthetic data. Future work will explore online and multi-modal extensions under dynamic distribution shifts.

\section{Acknowledgments}
This research was supported by the National Natural Science Foundation of China (Grants No. 72374154 and 62273261) and the Program of China Scholarship Council (Grant No. 202506260363).  
The work was carried out in part while the first author, Tianyi Yin, was a Visiting PhD Student at A*STAR, Singapore.
\bibliography{aaai2026}

\clearpage

\section*{Appendix: Experimental Details}

\subsection*{Forecasting Horizon and Input Window}
All experiments were conducted under a forecasting horizon of $H = 100$ time steps. The input sequence length was set to $L = 336$, which is a common setting in prior time-series forecasting studies. Each base model was initialized from officially released pretrained weights without any additional pretraining.

\subsection*{Dataset Splits}
For each dataset, we randomly split the entire time-series data into $80\%$ training and $20\%$ testing sets to validate the final performance. In the continual adaptation setting, we treat the five public datasets as old tasks and an in-house pedestrian flow dataset as the new task.

To simulate the few-shot scenario, we further subsample the new-task training set by randomly selecting $10\%$ of the training portion (i.e., $8\%$ of the full dataset) for fine-tuning. The remaining $72\%$ of the data in the training portion is discarded to mimic data-scarce adaptation.

\subsection*{Synthetic Replay Samples}
During adaptation, R-Tuning incorporates synthetic replay samples generated from the frozen pretrained model via Gaussian latent sampling. Unless otherwise specified, the replay size was fixed at $N = 2000$ per experiment. The redundant wavelet transform (RWT) was applied with a decomposition level of $L=1$, and the detail scaling factor was set to $\alpha = 0.7$.

\subsection*{Hyperparameter Settings}
The distillation temperature was fixed at $\tau = 3$. The weights for distillation loss and regularization were $\lambda = 0.2$ and $\beta = 10^{-4}$, respectively. All models were fine-tuned for $10$ epochs, and the checkpoint with the best validation MAE was selected for final evaluation.

\subsection*{Evaluation Metrics}
Model performance was evaluated using Mean Absolute Error (MAE) and Mean Squared Error (MSE), averaged across all prediction horizons. For old tasks, the metrics were computed by evaluating the adapted model on the entire old-task test sets. For new tasks, metrics were computed on the new-task test set, which corresponds to the held-out $20\%$ portion of the full dataset.

\subsection*{Computational Environment}
All experiments were conducted on cloud servers equipped with dual NVIDIA L40 GPUs (48 GB memory each). The implementation was based on Python 3.10, and all random seeds were fixed to ensure reproducibility.

\subsection*{Datasets}
Table~\ref{tab:dataset_stats} summarizes the key statistics of all datasets used in our experiments. 
For the five public datasets (ETT, ECL, Traffic, Weather, and Solar), we follow the standard preprocessing pipeline widely adopted in time-series forecasting benchmarks. 
The new-task dataset is an in-house pedestrian flow dataset collected from a commercial area in Shanghai. 

All datasets were normalized using z-score normalization:
\begin{equation}
    x' = \frac{x - \mu}{\sigma},
\end{equation}
where $\mu$ and $\sigma$ denote the mean and standard deviation computed from the training set. 

\begin{table}[h]
\centering
\setlength{\tabcolsep}{1mm}
\begin{tabular}{lccc}
\toprule
\textbf{Dataset} & \textbf{\#Variables} & \textbf{Frequency} & \textbf{Domain} \\
\midrule
ETT (ETTm1)  & 7   & 15 min & Energy \\
ECL          & 321 & 1 hour  & Electricity \\
Traffic      & 862 & 1 hour  & Traffic \\
Weather      & 21  & 10 min  & Meteorology \\
Solar        & 137 & 10 min  & Solar power \\
Crowd Flow   & 1   & 5 min   & Transportation \\
\bottomrule
\end{tabular}
\caption{Dataset statistics.}
\label{tab:dataset_stats}
\end{table}

\subsection*{Hyperparameters for Baseline Models}
Table~\ref{tab:baseline_hyperparams} lists the hyperparameter settings for each baseline model used in our experiments. 
All models were initialized with their officially released pretrained weights. 
For fine-tuning, we adjusted the learning rate and batch size based on validation performance, while other parameters remained consistent with their official configurations.

\begin{table}[h]
\centering
\setlength{\tabcolsep}{1mm}
\begin{tabular}{lccc}
\toprule
\textbf{Model} & \textbf{Learning Rate} & \textbf{Batch Size} & \textbf{Optimizer} \\
\midrule
Chronos    & $5\times 10^{-4}$ & 64 & AdamW \\
Apollo     & $5\times 10^{-4}$ & 64 & AdamW \\
Moirai     & $3\times 10^{-4}$ & 64 & Adam \\
AutoTimes  & $3\times 10^{-4}$ & 32 & AdamW \\
TimerXL    & $5\times 10^{-4}$ & 64 & AdamW \\
GPT4TS     & $2\times 10^{-4}$ & 32 & Adam \\
\bottomrule
\end{tabular}
\caption{Hyperparameter settings for baseline models.}
\label{tab:baseline_hyperparams}
\end{table}

\end{document}